\documentclass[final]{cvpr}

\usepackage{times}
\usepackage{epsfig}
\usepackage{graphicx}
\usepackage{amsmath}
\usepackage{amssymb}

\usepackage{booktabs}
\newcommand{\minisection}[1]{\vspace{0.04in} \noindent {\elvbf #1}\ \ }

\usepackage{color}
\definecolor{malachite}{rgb}{0.0, 0.45, 0.05}
\definecolor{mordantred19}{rgb}{0.68, 0.05, 0.0}

\usepackage[pagebackref=true,breaklinks=true,colorlinks,bookmarks=false]{hyperref}

\def\cvprPaperID{3426} 
\def\confYear{CVPR 2021}

\newcommand{\introfigure}{%
\begin{figure}[t]
\begin{center}
\includegraphics[width=0.42\textwidth]{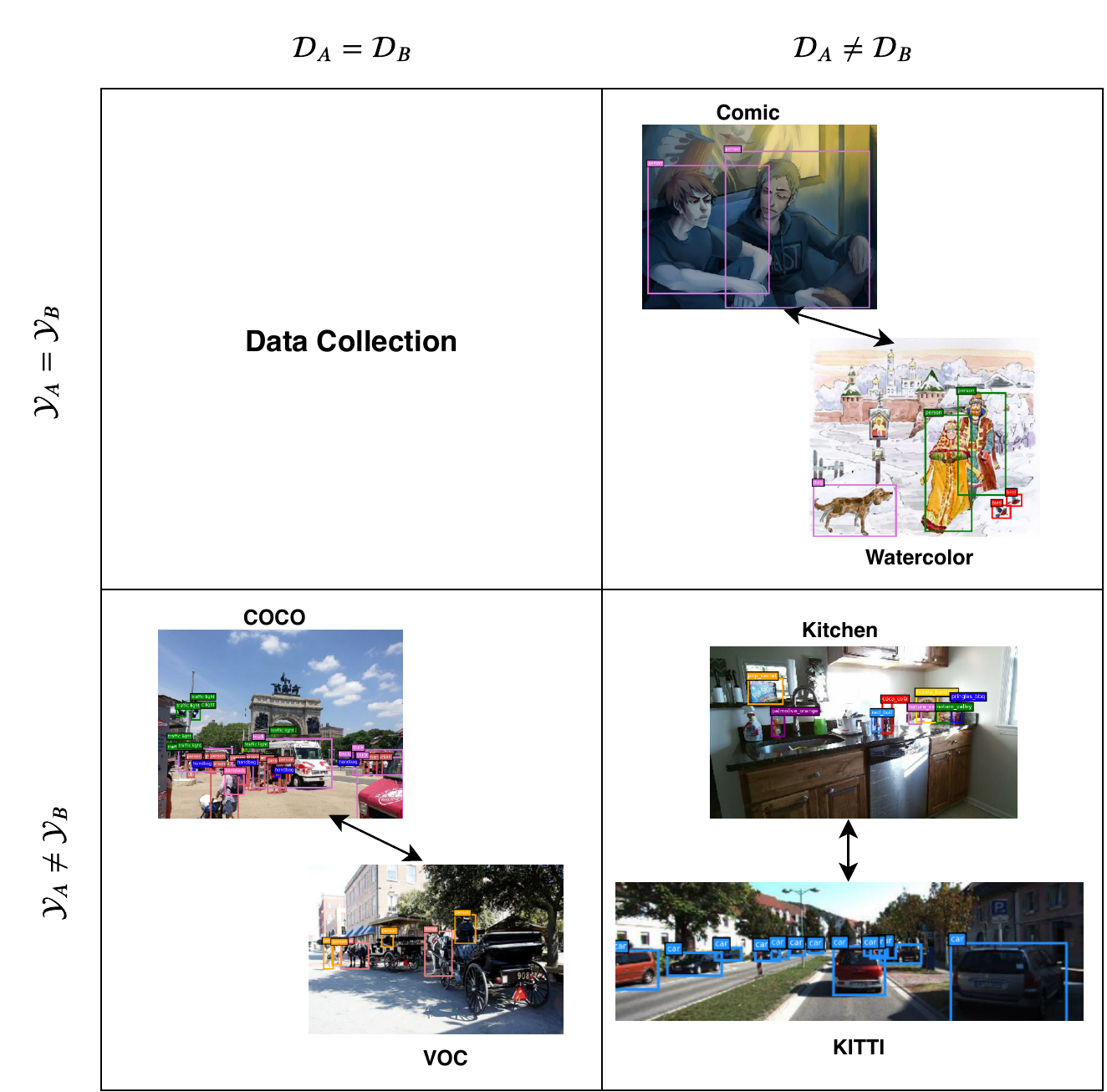}
\caption{Given $A$, $B$ two object detection tasks, $\mathcal{D}_{A}$, $\mathcal{D}_{B}$ their corresponding domains, and $\mathcal{Y}_{A}$, $\mathcal{Y}_{B}$ their corresponding categories, there are four different scenarios when learning $A$ and $B$ sequentially: more training data (top-left), same categories but different domains (top-right), different categories but same domains (bottom-left), and different categories and domains (bottom-right).} 
\label{fig:intro}
\end{center}
\vspace{-8 mm}
\end{figure}
}
\newcommand{\mainpipeline}{%
\begin{figure*}[t]
\begin{center}
\includegraphics[width=\textwidth]{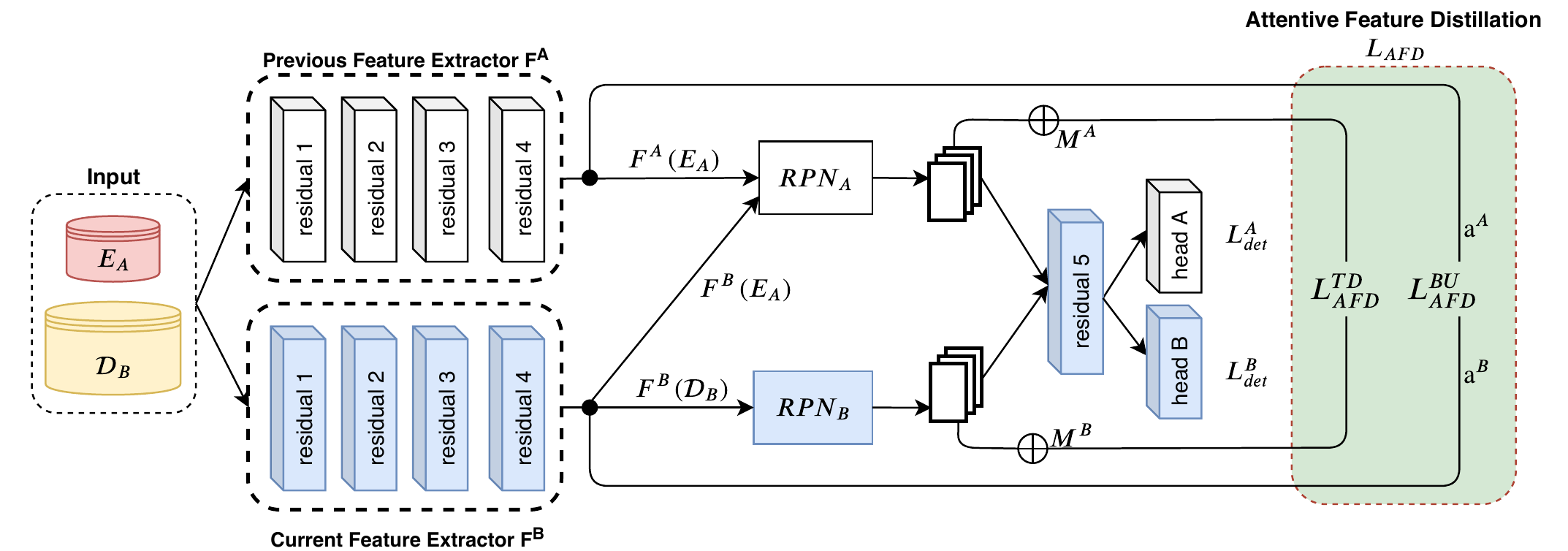}
\caption{The proposed framework for learning tasks $A,B$ sequentially based on Faster-RCNN (Best view in color). White modules on top are frozen after learning task $A$. When learning task $B$, we first duplicate the blue modules (including the RPN) and add a new head. Then, we combine data from task $B$ ($\mathcal{D_{B}}$) and exemplars from task $A$ ($E_{A}$). In green, our proposed approach combines bottom-up attention ($L_{\text{AFD}}^{\text{BU}}$) from the outputs of the feature extractors with top-down attention ($L_{\text{AFD}}^{\text{TD}}$) from the output bounding boxes of the RPN. The attentive feature distillation loss ($L_{\text{AFD}}$) is then combined with the usual object detection losses ($L_{\text{det}}^{A}$, $L_{\text{det}}^{B}$). The same procedure can be easily extended to more tasks.} 
\label{fig:framework}
\end{center}
\vspace{-3 mm}
\end{figure*}
}
\newcommand{\figattentionmaps}{%
\begin{figure}[tb]
\begin{center}
\includegraphics[width=0.99\linewidth]{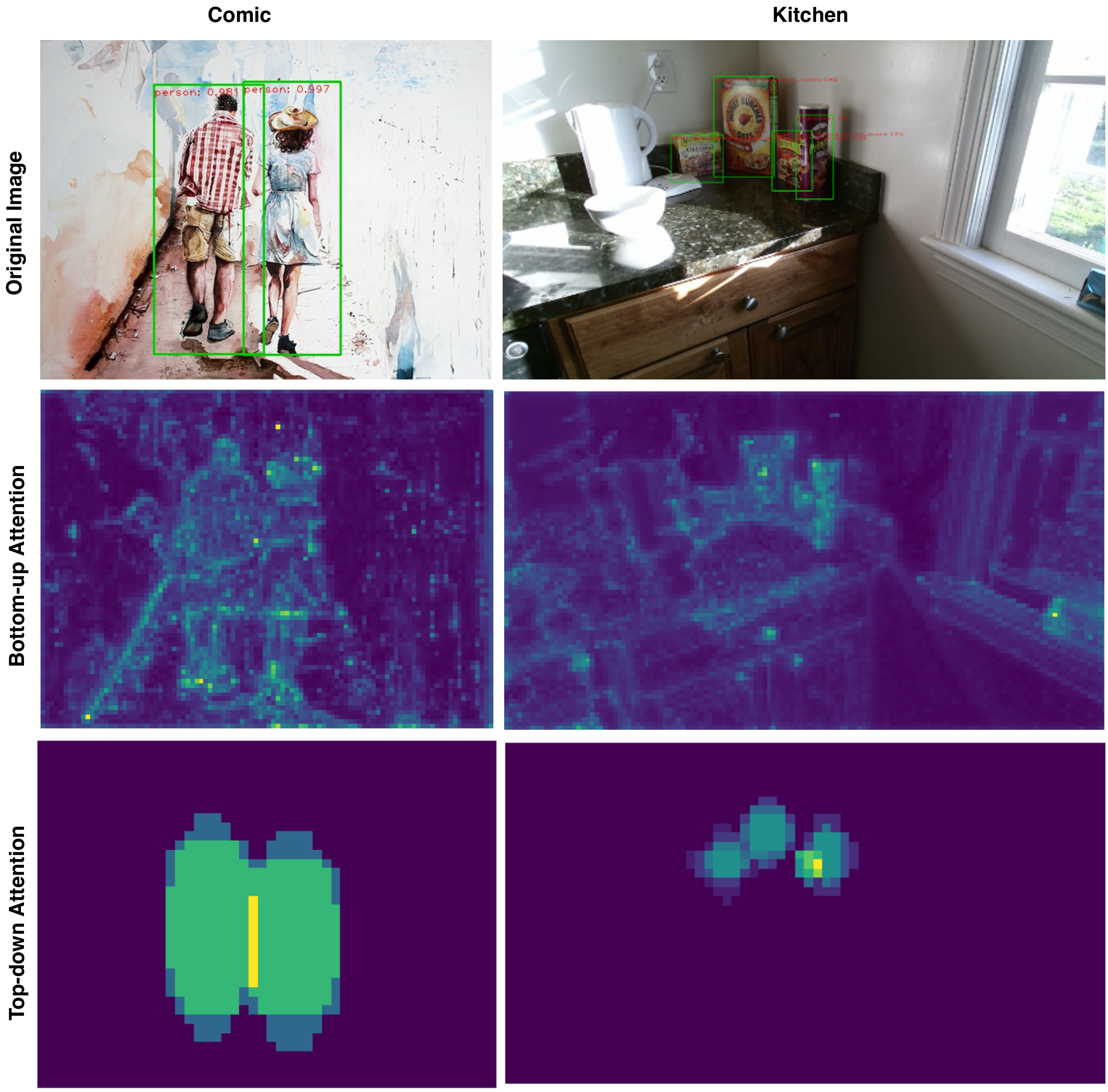}
\caption{Examples of attention maps. Original images (\textit{top}), bottom-up attention (\textit{middle}), top-down attention (\textit{bottom}) from Comic (\textit{left}) and Kitchen (\textit{right}) object detection datasets.} 
\label{fig:attention}
\end{center}
\vspace{-8 mm}
\end{figure}
}
\newcommand{\tabscenarios}{
\begin{table*}[tb]
\setlength{\tabcolsep}{4pt}
\caption{Analysis of performance (mAP) and forgetting on different scenarios without any exemplars. Arrows indicate order of learning. We evaluate on both datasets after training on the second dataset incrementally. }
\resizebox{0.99\textwidth}{!}{%
\begin{tabular}{l|cc|cc|cc|cc|cc|cc}
\hline
 & \multicolumn{4}{c|}{$\mathcal{D}_{A}\!\neq\!\mathcal{D}_{B}$ and $\mathcal{Y}_{A}\!\neq\!\mathcal{Y}_{B}$} & \multicolumn{4}{c|}{$\mathcal{D}_{A}\!\neq\!\mathcal{D}_{B}$ and $\mathcal{Y}_{A}\!=\!\mathcal{Y}_{B}$} & \multicolumn{4}{c}{$\mathcal{D}_{A}\!=\!\mathcal{D}_{B}$ and $\mathcal{Y}_{A}\!\neq\!\mathcal{Y}_{B}$} \\
\hline
Methods & \multicolumn{2}{c|}{KITTI $\rightarrow$ Kitchen} & \multicolumn{2}{c|}{Kitchen $\rightarrow$ KITTI} & \multicolumn{2}{c|}{Comic $\rightarrow$ Watercolor} & \multicolumn{2}{c|}{Watercolor $\rightarrow$ Comic} & \multicolumn{2}{c|}{VOC $\rightarrow$ COCO} & \multicolumn{2}{c}{COCO $\rightarrow$  VOC} \\
\hline
Joint Training &   53.7                  & 70.9    &   70.9        &   53.7   &       45.1        & 50.0           &    50.0          &  45.1      &         74.1          & 47.8     &    47.8      & 74.1  \\
\hline
Fine-tuning         &   10.9 (\textcolor{mordantred19}{-42.8})                  & 68.2     &   9.2 (\textcolor{mordantred19}{-61.7})       &   54.0    &       39.6 (\textcolor{mordantred19}{-5.5})          & 48.8           &    46.4 (\textcolor{mordantred19}{-3.6})           &43.4       &         74.9 (\textcolor{malachite}{+0.8})           & 48.4     &    20.1 (\textcolor{mordantred19}{-27.7})      & 75.9  \\
EWC~\cite{kirkpatrick2017overcoming}         &  15.8  (\textcolor{mordantred19}{-37.9})                  & 65.8    &   10.7 (\textcolor{mordantred19}{-60.2})       &  53.4    &       38.9 (\textcolor{mordantred19}{-6.2})          &      48.8    &   47.5  (\textcolor{mordantred19}{-2.5})           & 44.2     &     67.0     (\textcolor{mordantred19}{-7.1})           &   44.4   &  27.2    (\textcolor{mordantred19}{-20.6})    &  75.0  \\
MAS~\cite{aljundi2018memory}        &   8.9 (\textcolor{mordantred19}{-44.8})                  & 70.3    &   11.5 (\textcolor{mordantred19}{-59.4})       &   54.0    &       39.5 (\textcolor{mordantred19}{-5.6})          &      47.5     &    47.3 (\textcolor{mordantred19}{-2.7})           &  46.3    &   69.0 (\textcolor{mordantred19}{-5.1})           &  43.9   & 28.1 (\textcolor{mordantred19}{-19.7})      &  74.8 \\
AFD (bottom-up) &    25.2 (\textcolor{mordantred19}{-28.5})                 &   68.8      &       13.1 (\textcolor{mordantred19}{-57.8})                &  50.9     &   43.1 (\textcolor{mordantred19}{-2.0})           & 48.3          &        47.5 (\textcolor{mordantred19}{-2.5})                  &  43.5       &     75.4 (\textcolor{malachite}{+1.3})  & 45.1      &   23.1 (\textcolor{mordantred19}{-24.7})                &  76.2   \\
AFD (top-down)  &     33.9 (\textcolor{mordantred19}{-19.8})               &  72.1       &    16.8 (\textcolor{mordantred19}{-54.1})          &    51.2   &        43.6 (\textcolor{mordantred19}{-1.5})              & 45.9           &     51.3 (\textcolor{malachite}{+1.3})         &46.2       &          76.0 (\textcolor{malachite}{+1.9})     &  45.5  &     28.1 (\textcolor{mordantred19}{-19.7})               &  76.8   \\
AFD (both) &     36.6 (\textcolor{mordantred19}{-17.1})              &    72.0     &     20.5 (\textcolor{mordantred19}{-50.4})                  &  50.5     &      42.9 (\textcolor{mordantred19}{-2.2})               &    48.5    &           49.4 (\textcolor{mordantred19}{-0.6})        &45.5       &    75.4 (\textcolor{malachite}{+1.3})       &    45.1  &     27.8 (\textcolor{mordantred19}{-20.0})       & 77.1    \\ \hline
\end{tabular}}
\label{tab:scenario}
\end{table*}
}

\newcommand{\tabsampling}{
\begin{table}[tb]
\setlength{\tabcolsep}{4pt}
\caption{Ablation study on different sampling methods (100 exemplars in total). }
\centering
\resizebox{0.95\linewidth}{!}{%
\begin{tabular}{l|c|cc|cc|c}
\hline
method & samples & \multicolumn{2}{c|}{Kitchen $\rightarrow$ KITTI} & \multicolumn{2}{c|}{KITTI $\rightarrow$ Kitchen} & Average \\
\hline
none                    & 0   & 20.5 & 50.5 & 36.6 & 72.0 & 44.9\\
random                  & 100 & 66.3 & 53.8 & 46.7 & 73.1 & 60.0\\
hard                    & 100 & 69.4 & 51.6 & 42.1 & 71.7 & 58.7\\
adaptive ($\eta\!=\!3$) & 100 & 68.1 & 52.5 & 45.9 & 71.4 & 59.5\\
adaptive ($\eta\!=\!5$) & 100 & 68.6 & 53.4 & 48.4 & 72.0 & \textbf{60.6}\\
adaptive ($\eta\!=\!7$) & 100 & 67.8 & 52.6 & 48.1 & 72.4 & 60.2\\
\hline
\end{tabular}}
\label{tab:sampling}
\end{table}
}

\newcommand{\figexemplars}{
\begin{figure}[tpb]
\centering
\includegraphics[width=0.49\linewidth]{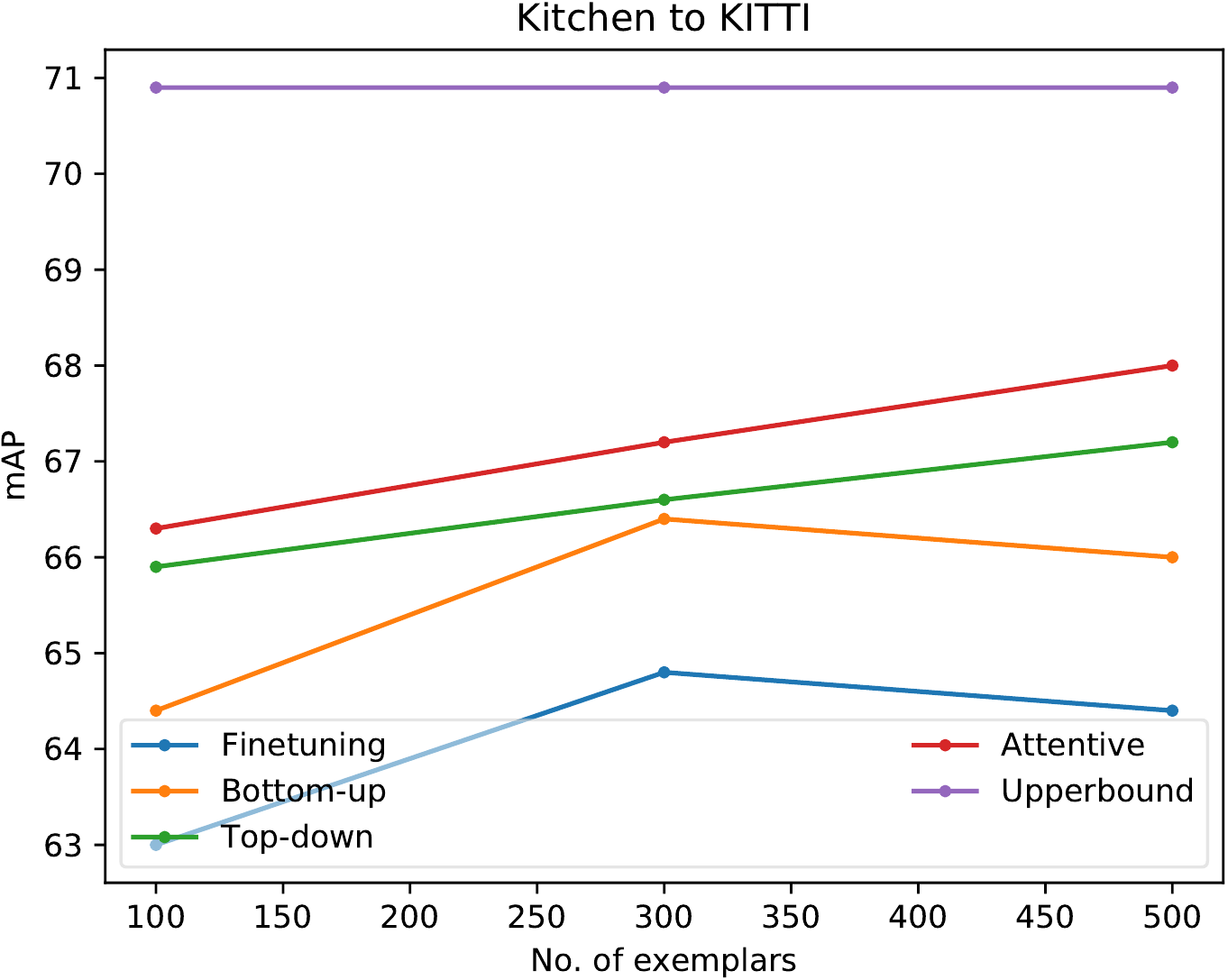} 
\includegraphics[width=0.49\linewidth]{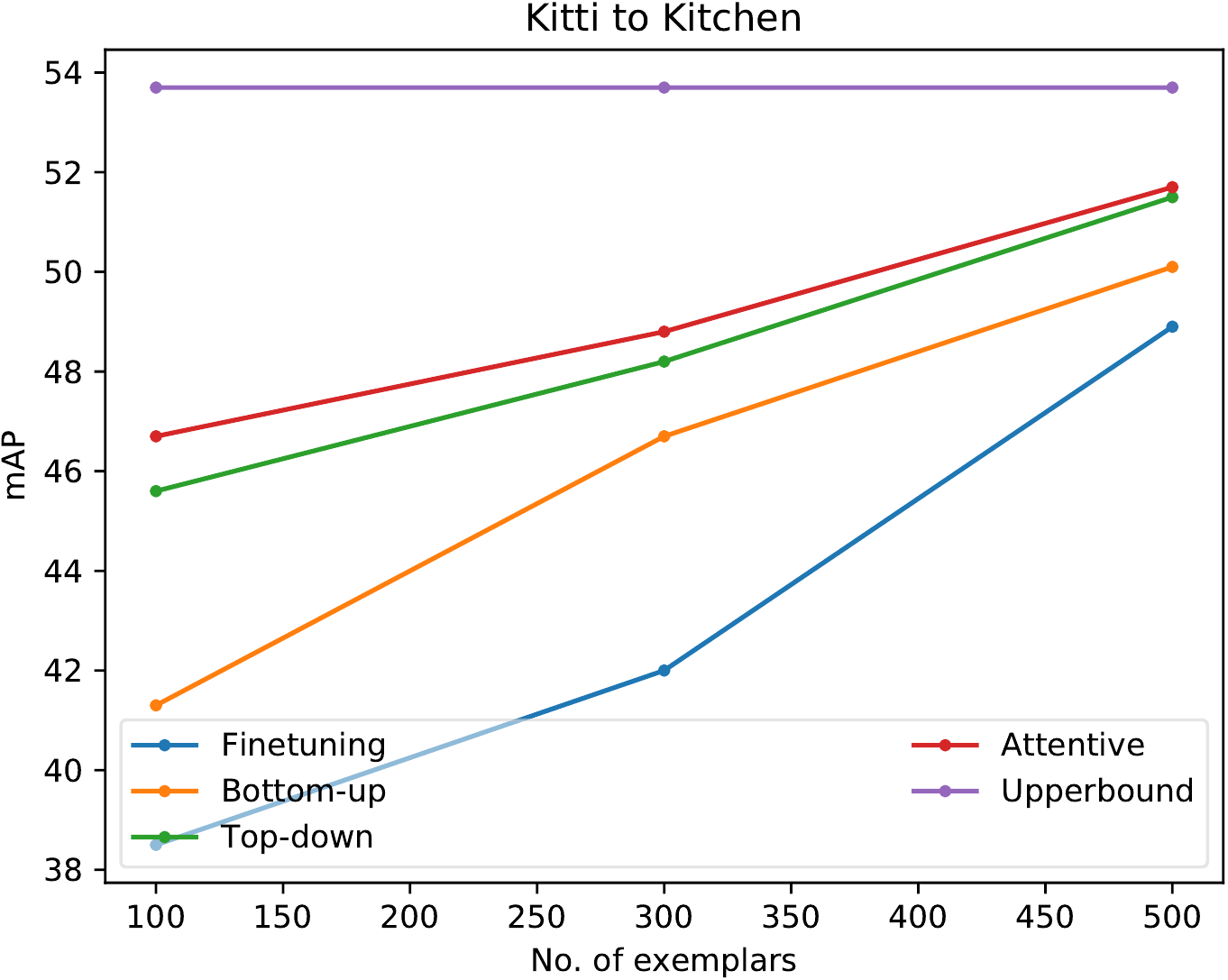}
\caption{Ablation study on different number of exemplars. Evaluation on the first dataset after training on the second dataset.}
\label{fig:exemplar}
\vspace{-6 mm}
\end{figure}
}

\newcommand{\figvisualization}{
\begin{figure*}[tb]
\begin{center}
\includegraphics[width=\textwidth]{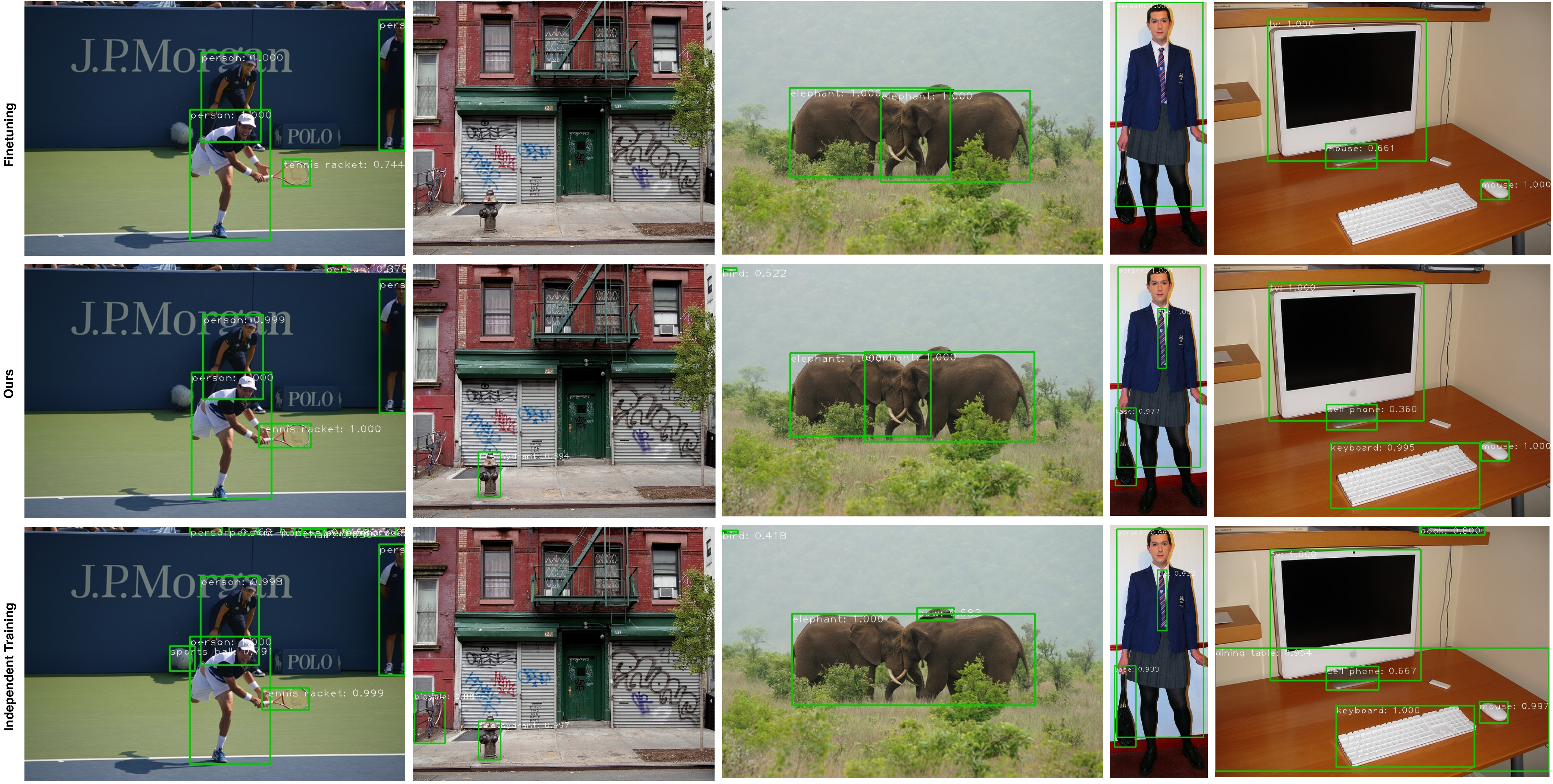}
\caption{Examples of predictions on COCO after training on VOC for Fine-tuning* (top), Joint Training (bottom) and Ours* (middle). Finetuning* tends to forget objects, such as the hydrant, the tie or the keyboard. Ours* is capable of detecting all those objects and only misses the ball and the bicycle. Joint Training is capable of detecting all the objects missed by the other methods, but is not capable of differentiating between the two elephants. In the last column, Ours* predicts the objects with higher confidence than the other methods.} 
\label{fig:coco}
\end{center}
\end{figure*}
}

\newcommand{\tabkittikitchen}{
\begin{table}[tb]
\setlength{\tabcolsep}{4pt}
\caption{Results on KITTI and Kitchen (100 exemplars in total). The mAP is evaluated on both datasets after training on the second dataset. Arrows indicate order of learning. ‘Average' means the average of numbers in the same row for easy comparison. ‘*' indicates the methods with exemplars.}
\centering
\resizebox{0.98\linewidth}{!}{%
\begin{tabular}{l|cc|cc|c}
\hline
Methods & \multicolumn{2}{c|}{Kitchen $\rightarrow$ KITTI} & \multicolumn{2}{c|}{KITTI $\rightarrow$ Kitchen} & Average\\
\hline

Joint Training            & 74.0 & 54.3 & 54.3 & 74.0 & 64.1\\ \hline
Fine-tuning                              & 9.2 & 54.0 & 10.9 & 68.2 & 35.6 \\
Fine-tuning*                            & 63.0 & 54.5 & 38.5 & 70.5 & 56.6\\
LwF Detection*~\cite{shmelkov2017incremental} & 59.9 & 54.7 & 39.4 & 69.9 & 56.0\\ 
Feature Distillation*~\cite{romero2014fitnets}                    & 62.7 & 54.4 & 35.0 & 69.4 & 55.4\\
Attention Distillation*~\cite{zagoruyko2016paying}                  & 64.2 & 52.8 & 39.8 & 71.0 & 57.0\\

EWC*~\cite{kirkpatrick2017overcoming}          & 68.4 & 52.8 & 48.3 & 65.5  & 58.8 \\
MAS*~\cite{aljundi2018memory}          & 67.7 & 55.6 & 42.8  & 71.7  & 59.5\\
AFD* (Ours)                                    & 68.6 & 53.4 & 48.1 & 72.4 & \textbf{60.6}\\
\hline
\end{tabular}}
\label{tab:kit}
\end{table}
}

\newcommand{\tabcomicwatercolor}{
\begin{table}[tb]
\setlength{\tabcolsep}{4pt}
\caption{Results on Comic and Watercolor (without exemplars). }
\centering
\resizebox{0.98\linewidth}{!}{%
\begin{tabular}{l|c|c|c}
\hline
Methods & Comic $\rightarrow$ Watercolor & Watercolor $\rightarrow$ Comic & Average\\
\hline

Joint Training & 45.3 \qquad 49.7 & 49.7 \qquad 45.3 &47.5 \\ \hline
Fine-tuning                  & 39.6 \qquad 48.8 & 46.4 \qquad 43.4 & 44.6 \\
LwF Detection~\cite{shmelkov2017incremental} & 39.8 \qquad 48.3 & 44.2 \qquad 44.7 & 44.3\\
Feature Distillation~\cite{romero2014fitnets}         & 39.7 \qquad  48.5 & 43.8 \qquad 41.0 & 43.3\\
Attention Distillation~\cite{zagoruyko2016paying}       & 40.9 \qquad 47.9 & 48.1 \qquad 44.6 & 45.4\\
EWC~\cite{kirkpatrick2017overcoming}       & 38.9 \qquad 48.8 & 47.5 \qquad 44.2 & 44.9\\
MAS~\cite{aljundi2018memory}       & 39.5 \qquad 47.5 & 47.3 \qquad 46.3 & 45.2\\
AFD (Ours)                         & 42.9 \qquad 48.5 & 49.4 \qquad 45.5 & \textbf{46.6}\\
\hline
\end{tabular}}
\label{tab:comicw}
\end{table}
}

\newcommand{\tabcocovoc}{
\begin{table}[tb]
\setlength{\tabcolsep}{4pt}
\caption{Results from COCO to VOC (500 exemplars in total). ‘*' indicates the methods with exemplars.}
\centering
\resizebox{0.68\linewidth}{!}{%
\begin{tabular}{l|c|c}
\hline
Methods & COCO $\rightarrow$ VOC & Average \\
\hline

Joint Training & 44.3 \qquad 79.0 & 61.7\\ \hline
Fine-tuning    & 20.1 \qquad 75.9 & 48.0 \\
Fine-tuning*      & 28.7 \qquad 73.3 & 51.0 \\
LwF Detection*~\cite{shmelkov2017incremental} & 26.6 \qquad 73.0 & 49.8\\
Feature Distillation*~\cite{romero2014fitnets}           & 26.9 \qquad 72.4 & 49.7 \\
Attention Distillation*~\cite{zagoruyko2016paying}         & 28.5 \qquad 73.0 & 50.8\\
EWC*~\cite{kirkpatrick2017overcoming}         &32.2  \qquad 73.4  & 52.8 \\
MAS*~\cite{aljundi2018memory}         & 32.7 \qquad 73.4 & 53.1\\

AFD* (Ours)                           & 36.8 \qquad 75.2 & \textbf{56.0}\\
\hline
\end{tabular}}
\label{tab:voc}
\end{table}
}

\newcommand{\tabthree}{
\begin{table}[tb]
\setlength{\tabcolsep}{4pt}
\caption{Results on KITTI, VOC and Kitchen (300 exemplars per task). The Pascal mAP is reported on all datasets after learning on the last. ‘*' indicates the methods with exemplars.}
\centering
\resizebox{\linewidth}{!}{%
\begin{tabular}{l|c|c|c}
\hline
 & KITTI $\rightarrow$ VOC $\rightarrow$ Kitchen & Kitchen $\rightarrow$ VOC $\rightarrow$ KITTI & Average\\
\hline
Fine-tuning* & 43.8 \qquad 55.0 \qquad 68.8 & 64.6 \qquad 58.4 \qquad 57.5 & 58.0 \\
AFD*  (Ours)      & 53.2 \qquad 63.4 \qquad 71.2 & 69.1 \qquad 65.0 \qquad 59.6 & \textbf{63.6}\\
\hline
\end{tabular}}
\label{tab:three}
\end{table}
}

\newcommand{\figforgetting}{
\begin{figure}[ht]
\centering
\includegraphics[width=0.99\linewidth]{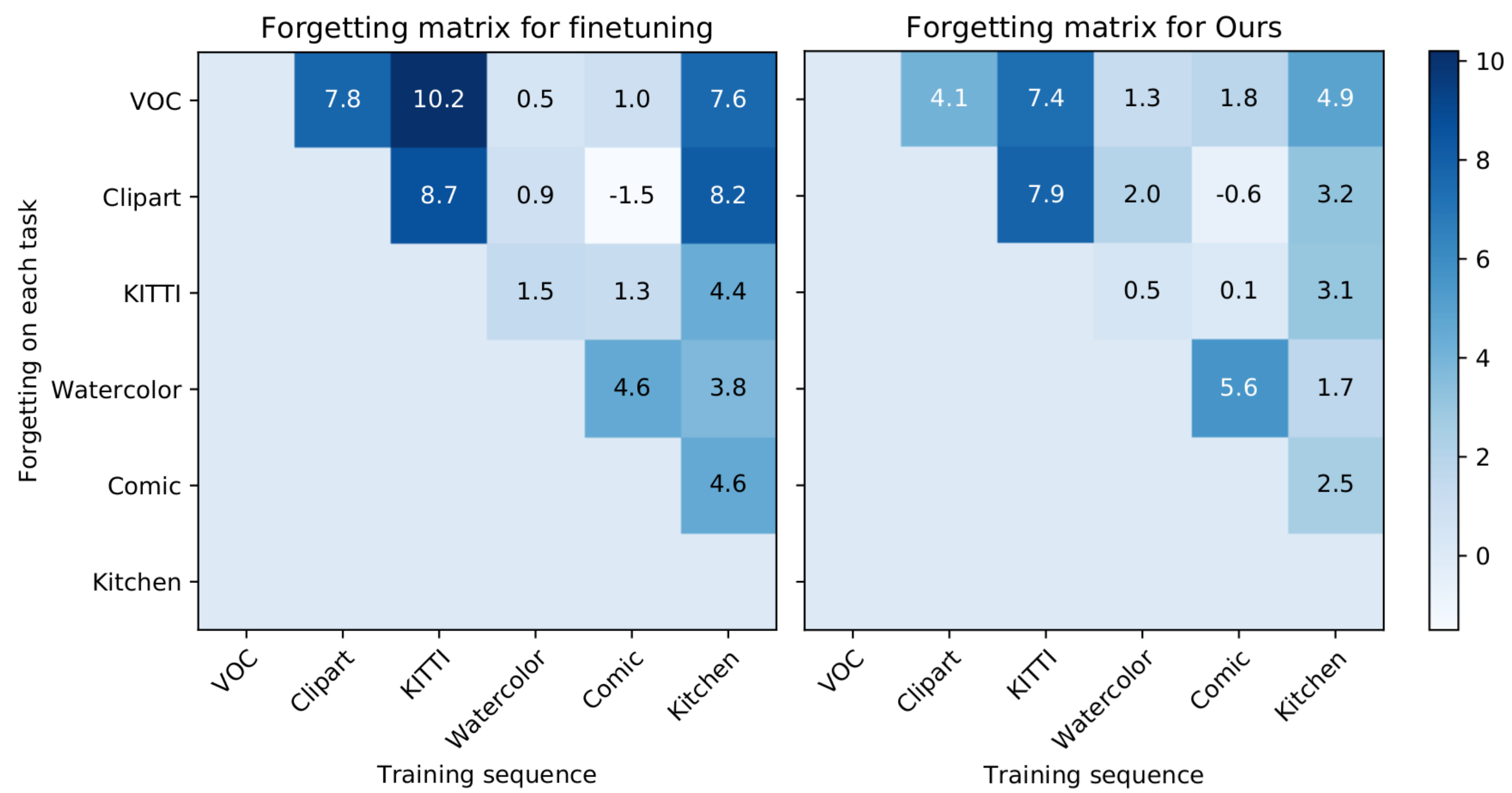} 

\caption{Forgetting matrix for Fine-tuning* and AFD* (right) on sequential learning of six tasks (300 exemplars per task). Horizontal axis represents the sequential learning order of different tasks. Vertical axis is forgetting mAP for each task after learning on the task in the horizontal axis. Each Row represents forgetting progression of each dataset over the learning sequence. The lighter the color, the less the forgetting.  }
\label{fig:six}
\end{figure}
}

\begin{document}

\title{Multi-Task Incremental Learning for Object Detection}

\author{Xialei Liu$^{1}$\thanks{Work done as an intern at Amazon Web Sevices.} ,  Hao Yang$^{2}$, Avinash Ravichandran$^{2}$, Rahul Bhotika$^{2}$, Stefano Soatto$^{2,3}$\\
$^{1}$ Computer Vision Center, UAB,  $^{2}$ Amazon Web Services, $^{3}$ UCLA\\
{\tt\small xialei@cvc.uab.es, \{haoyng,ravinash,bhotikar,soattos\}@amazon.com}
}

\maketitle

\begin{abstract}
   Multi-task learns multiple tasks, while sharing knowledge and computation among them. However, it suffers from catastrophic forgetting of previous knowledge when learned incrementally without access to the old data. Most existing object detectors are domain-specific and static, while some are learned incrementally but only within a single domain. Training an object detector incrementally across various domains has rarely been explored. In this work, we propose three incremental learning scenarios across various domains and categories for object detection. To mitigate catastrophic forgetting, attentive feature distillation is proposed to leverages both bottom-up and top-down attentions to extract important information for distillation. We then systematically analyze the proposed distillation method in different scenarios. We find out that, contrary to common understanding, domain gaps have  smaller negative impact on incremental detection, while category differences are problematic. For the difficult cases, where the domain gaps and especially category differences are large, we explore three different exemplar sampling methods and show the proposed adaptive sampling method is effective to select diverse and informative samples from entire datasets, to further prevent forgetting. Experimental results show that we achieve the  significant improvement in three different scenarios across seven object detection benchmark datasets.
\end{abstract}

\section{Introduction}

Object detection has improved significantly in recent years~\cite{cai2018cascade,girshick2015fast,lin2017feature,ren2015faster} on challenging benchmarks, such as PASCAL VOC~\cite{everingham2010pascal} and MS COCO~\cite{lin2014microsoft}. However, most existing detectors are domain-specific, lack generalization ability to new domains and categories. Recently, Wang et al.~\cite{wang2019towards} proposed a multi-task universal object detector by domain attention across eleven diverse datasets. They show that a single network with marginally more parameters can handle detection on multiple domains, even outperforming single-domain detectors in most cases.
However, a major drawback of such a multi-task universal detector is that all the data has to be accessible during the  model training. Moreover, once the model is trained, it is hard to adapt it to new domains without retraining all seen domains again. Unfortunately, in real-world scenarios, legacy data from previous domains can be lost, proprietary, or simply too expensive to use in new domains~\cite{li2017learning}. For example, for models deployed on mobile devices, it is unlikely that they can still access the original training data. Even if they do have access, the limited resources on mobile devices make it difficult to retrain with full training data. Uploading the new training data to the cloud might not be an option due to privacy concerns. If we want to learn to detect/localize new concepts from new domains on devices for these models \textit{without losing the detection ability in previous domains}, we need to develop a new learning scheme. 

\introfigure  

Multi-Task incremental learning  learns new knowledge by updating a model trained from previous datasets incrementally. It is well analyzed in image classification~\cite{de2019continual,parisi2019continual}, while it has rarely been explored in object detection. Most work focus on incremental object detection in a single domain~\cite{joseph2020incremental,shmelkov2017incremental,zhou2020lifelong}. In this paper, we  propose  three scenarios to learn an object detector across multiple datasets incrementally varying categories and domains, as shown in Figure~\ref{fig:intro}. For instance, Watercolor and Comic~\cite{inoue2018cross} share the same categories, but they are from different domains, how would domain shift affect the continual learning process? VOC and COCO are from similar domains, but COCO has 60 more categories, how would category difference affect? KITTI~\cite{geiger2012we} and Kitchen~\cite{georgakis2016multiview} have disjoint categories and significant domain discrepancies, how  difficult is it to learn sequentially? To answer these questions and better understand forgetting across domains, we conducted a comprehensive analysis in different scenarios. 

Distillation is widely used in image classification to avoid \textit{catastrophic forgetting}~\cite{achille2018life,dhar2019learning,li2017learning,romero2014fitnets}. It is ineffective to apply them directly to object detection, since in state-of-the-art two-stage and some one-stage detectors~\cite{yolov3,ren2015faster},  majority of box hypotheses (``proposals'') are background (provided by either regional proposal networks or dense sampling). Due to complexity of background and various sizes/numbers of foreground objects, treating the whole image and all region proposals equally will lead to relatively weak guidance from the important foreground objects. Therefore, we need to have top-down attention from the ground-truth bounding boxes to guide the distillation process. Also, it has been shown in ~\cite{context,wang2019distilling} that context information is very important for object detection. Thus, we cannot discard all background information and need to preserve useful background information and context around foreground objects. Therefore, we propose attentive feature distillation to apply both top-down attention and bottom-up attention on feature maps for object detection.

 We show in our experiments that category differences cause a bigger problem of forgetting, and just using attentive feature distillation is not enough to prevent forgetting in these difficult cases. Storing a small number of exemplars from previous tasks is commonly used in classification, but it is unclear how it would affect in object detection. Therefore, we explore three different ways to use of a small number of exemplars from previous tasks to further mitigate forgetting. Especially, our proposed adaptive sampling method can balance between the number of bounding boxes and the diversity of exemplars.  
 
 To summarize, we make the following contributions:
\begin{itemize}
  
  \item We propose three scenarios across multiple datasets and analyze the effects of domain gaps and category differences on catastrophic forgetting for incremental object detection. 
    
  \item We propose an attentive feature distillation approach to distill important knowledge using both top-down and bottom-up attentions. 

  \item We explore three different exemplar sampling methods for object detection and propose an adaptive exemplar sampling to choose informative and diverse samples for difficult cases.
\end{itemize}
\noindent Experimental results across seven datasets demonstrate the effectiveness of our proposed method and show interesting behaviors in different scenarios.

\section{Related Work}
\label{sec:related}

\minisection{Object Detection} Object detection networks can be divided into two categories: one and two-stage. Faster-RCNN~\cite{ren2015faster} is a representative two-stage method with a region proposal network (RPN) and a classification and regression network (Fast-RCNN~\cite{girshick2015fast}). YOLO~\cite{redmon2016you} and SSD~\cite{liu2016ssd} are representatives of one-stage methods, which predict bounding boxes and class probabilities directly from full images in one step. Object detection has achieved significant improvement on multiple large datasets, such as PASCAL VOC~\cite{everingham2010pascal}, MSCOCO~\cite{lin2014microsoft}. However, most of the existing detectors are domain and category specific. Recently, Wang et al.~\cite{wang2019towards} proposed to train a single network across datasets jointly. In this work, the network is learned incrementally and dynamically across multiple datasets.

\minisection{Incremental Learning}
 Exiting work on incremental learning mostly focus on classification problems, which can be roughly divided into three main families~\cite{de2019continual,parisi2019continual}: regularization-based,  parameter-isolated and rehearsal-based methods. Regularization-based methods contain both data-focused and prior-focused approaches. Data-focused methods are composed of distillation approaches such as LwF~\cite{li2017learning} using probability distillation and LwM~\cite{dhar2019learning} using attention distillation. Prior-focused methods estimate the importance of parameters, and it applies a higher penalty on those that show a significant change when switching from one task to another~\cite{aljundi2018memory,kirkpatrick2017overcoming,li2017learning,liu2018rotate,zenke2017continual}. The next family prevents catastrophic forgetting by growing a sub-network or learning a mask for each task~\cite{lee2017lifelong,mallya2018piggyback,mallya2018packnet,rusu2016progressive,serra2018overcoming}. Finally, rehearsal-based approaches form the last family, which either store a small number of training samples from previous tasks~\cite{chaudhry2018riemannian,lopez2017gradient,rebuffi2017icarl}, or use a generative model to sample synthetic data from previously learned distributions~\cite{shin2017continual,wu2018memory}.  

There are relatively few methods for incremental object detection learning. 
Shmelkov et al.~\cite{shmelkov2017incremental} proposed the first method to learn an object detector incrementally by adding one or a few classes at a time within one single domain. It leverages the distillation loss for both object localization and object classification. Distillation-based methods are further improved by~\cite{hao2019end,Ramakrishnan_2020_CVPR_Workshops,yang2020two,zhou2020lifelong}. However, they still suffer from noisy backgrounds and noisy region proposals. Meta-learning was proposed for incremental object detection to reshape model gradients for optimally sharing information~\cite{joseph2020incremental}. All these methods are applied to incremental object detection in a single domain.  In this work, we propose three scenarios across multiple datasets. Recently, It has also been studied on few-shot incremental object detection~\cite{Perez-Rua_2020_CVPR,rahman2020any} and online object detection~\cite{acharya2020rodeo}. 

\minisection{Transfer Learning}  Convolutional neural networks (CNNs) models trained on ImageNet have been widely used for transfer learning; either using a pre-trained model as a feature extractor\cite{donahue2014decaf} or fine-tuning the model on the target data\cite{girshick2014rich,oquab2014learning}. After transfer learning,  performance on target task is only considered important and performance on the original task is ignored. We refer to~\cite{pan2009survey} for a comprehensive survey on transfer learning. In this paper, we  transfer knowledge of the object detector across domains incrementally but aim at avoiding the catastrophic forgetting of previous domains. Since there is no prior work on analyzing the relation between the forgetting and domain-category differences, we conduct in-depth forgetting analysis during transfer learning.

\section{Preliminaries} 
\label{sec:pre}
In this section, we introduce the notation for object detection in multi-task incremental learning and present the proposed approach in Sec.~\ref{sec:pro}.

\subsection{Object Detection}
An object detector is a function that maps an image $\mathbf{x} \in {\mathcal X}$ to an output $\mathbf{y} \in {\cal Y}$ which consists of multiple bounding box and their labels. Object detectors are typically trained from a dataset $\left\{\left(\mathbf{x}_i,\mathbf{y}_i\right)\right\}_{i=1}^{N}$ of size $N$, by minimizing a loss or objective $L$. In the case of a typical two-stage detector, the function is composed of three elements, a {\em feature extractor} $F$ that maps a (region of the) image onto a vector or probability space,  $\mathbf{z}=F(\mathbf{x})$, a {\em region proposal network} (RPN), $R$ that samples hypotheses $\hat{\mathbf{y}} = R(\mathbf{z})$,
and a classifier, or {\em head} $H$ that measures the fit of the hypothesis, $H(\hat{\mathbf{y}}, \mathbf{y})$. 

\begin{equation}
   L (F,R,H)=-\frac { 1 }{ N } \sum _{ i=1 }^{ N }{ { L }_{\text{det}}(H(R(F({ \mathbf{x} }_{ i }))), { \mathbf{y} }_{ i }) },
\end{equation}
where detection loss ${ L }_{\text{det}}$ accounts for both label classification error  and bounding box(es) regression error. $L_{\text{det}} = L_{reg} + L_{cls}$, where $L_{cls}$ and $L_{reg}$ are defined as cross entropy loss and smooth-$L_1$ loss, respectively. The functions $F, R, H$ are typically implemented using deep neural networks (DNNs), and training or learning corresponds to minimizing the loss with respect to their parameters.  

\subsection{Incremental Multi-Task Object Detection}
\label{sec:continualobject}
In continual/incremental learning, $T$ tasks are presented sequentially, each corresponding to a different dataset of {\em domains} $\mathcal{D}_1,\ldots,\mathcal{D}_T$ and categories $\mathcal{Y}_1,\ldots,\mathcal{Y}_T$. Since training $T$ models independently is inefficient both statistically and computationally, in typical incremental/continual learning setting, parts of the model are shared across all tasks, most notably the feature extractor $F$. Then, multi-task incremental learning methods focus on solving the catastrophic forgetting problem when training the tasks sequentially with shared backbone features and separate heads~\cite{de2019continual}. Due to the complexity of background and diversity in sizes and numbers of foreground objects, compared to incremental classification, incremental detection needs more specially designed to mitigate forgetting, which we will introduce in Section~\ref{sec:pro}. Without loss of generality, we present and analyze our approach for the case of two-task incremental detection. Extensions to more tasks is straightforward and we validate this claim by showing experiments on more than two tasks in Section~\ref{sec:moretasks}.

\section{Proposed Approach}
\label{sec:pro}
\mainpipeline

 We decompose two-task incremental learning into four different scenarios for in-depth understanding of how domain gaps and category differences contribute to catastrophic forgetting. Then we describe our overall framework on multi-task incremental detection. It is followed by attentive feature distillation that uses both bottom-up and top-down attentions, and proposed exemplar sampling methods.

\subsection{Three Scenarios for Incremental Detection}
\label{sec:fourscenarios}
We consider four different scenarios (three non-trivial scenarios) in two-task incremental detection, using the two tasks $A$ and $B$ for reference (Fig.~\ref{fig:intro}): (i) both domains and classes are the same,  so there is no covariate shift; (ii) Same domain, different categories, $\mathcal{D}_{A} = \mathcal{D}_{B}$ and $\mathcal{Y}_{A} \neq  \mathcal{Y}_{B}$, corresponding to intra-domain sequential learning. (iii) Domain shift, $\mathcal{D}_{A} \neq \mathcal{D}_{B}$ but $\mathcal{Y}_{A} = \mathcal{Y}_{B}$, corresponding to domain adaptation, say from real to synthetic images. Finally (iv) both domains and categories differ, $\mathcal{D}_{A} \neq \mathcal{D}_{B}$ and $\mathcal{Y}_{A} \neq \mathcal{Y}_{B}$. For example, KITTI and Kitchen datasets have different domain (indoor vs. outdoor imagery) and categories. In the following sections, we focus on the last three non-trivial scenarios.


\subsection{Framework}
Our incremental detection system is based on a generic two-stage object detection algorithms (Faster-RCNN~\cite{ren2015faster}, FPN~\cite{lin2017feature}). We assume ResNet-style networks (ResNet~\cite{he2016resnet}, SE-ResNet~\cite{seresnet}) as backbone.
Specifically, for the two-task case, as shown in Figure~\ref{fig:framework}, we train the first task $A$ using Faster-RCNN with data $\mathcal{D}_{A}$. Then we train on task $B$ with data $\mathcal{D}_{B}$ using our proposed attentive feature distillation (Section~\ref{sec:afd_proposal}) and adaptive exemplar sampling method (Section~\ref{sec:sampling}). After learning on the current task, only current model is needed to do inference for all tasks and previous model is removed from the memory. This framework could be easily extend to more tasks.

By fusing the usual object detection loss functions with the proposed attentive feature distillation loss, the final objective function for incremental object detection learning from task A to task B is defined as:

\begin{equation}
L(F^{B}, R^{B}, H^{B}) = { L }_{det}^{B} + { L }_{det}^{A} + { \lambda L }_{ AFD },
\label{eq:overall}
\end{equation}

\noindent where ${ L }_{det}^{B}$ and ${ L }_{det}^{A}$ are the standard Faster-RCNN detection loss, which is updated using data from current task B and a few exemplars (in Section~\ref{sec:sampling}) selected from previous task A, respectively. ${L}_{AFD}$ is our propsoed attentive feature distillation loss and we will explain it in the next section. $\lambda$ is a trade-off parameter to balance the current learning objective and distillation objective to avoid forgetting.

\subsection{Attentive Feature Distillation (AFD)}
\label{sec:afd_proposal}

After learning on the first task $A$, we train on task $B$ incrementally, the bottom-up AFD is defined as integrating self-attention maps and feature maps. Here self-attention maps are summation of all channels of the feature maps. They capture the important blobs in the feature maps based on network itself. By combining self-attention maps and feature maps, we can capture the important context and background information smoothly and naturally during training, as shown in the middle row in Figure~\ref{fig:attention}. 
Formally, we define bottom-up AFD as:
\begin{equation}
\label{eq:bottomup}
    { L }^{\text{BU}}_{ \text{AFD} } (F^{B}) =  \frac { 1 }{ 2 } { \left\| \frac { { \text{a} }_{ i }^{ B } }{ \left\| { \text{a} }_{ i }^{ B } \right\|  } { F }^{ B }({ \mathbf{x} }_{ i })-\frac { { \text{a} }_{ i }^{ A } }{ \left\| { \text{a} }_{ i }^{ A } \right\|  } { F }^{ A }({ \mathbf{x} }_{ i }) \right\|  }^{ 2 }.
\end{equation}
Here $F^{B}$  and $F^{A}$ are the feature extractors of current task $B$ and previous task $A$ (Fig.~\ref{fig:framework}) and the self-attention maps are
\begin{equation}
\label{eq:att}
    { \text{a} }_{ i }=\sum _{ j=1 }^{ c }{ { \left| { { F }  } ({ \mathbf{x}_{ i } })^{ j } \right|  }^{ 2 } },
\end{equation}
where $c$ is the number of channels per activation.

\figattentionmaps  

Top-down attention is computed using the ground truth information. For each ground truth bounding box $G_{i}$, intersection-over-union (IoU) $G_{i} \cup P_{j}$ is computed for each box proposal $P_{j}$ provided by the RPN. The overlapping region between $G_{i}$ and $P_{j}$ is defined as $O_{i,j}$:

\begin{equation}
    O_{i,j}= 
\begin{cases}
    1,& \text{if } G_{i} \cup P_{j} > th \\
    0,              & \text{otherwise}
\end{cases}
,
\end{equation}
A threshold $th$ is set as $\frac{1}{2} \max{\left(G_{i} \cup \sum_{j=1}^{\left | P \right |} P_{j}\right)}$ to decide the correct predictions. $\left | P \right |$ is the number of region proposals. The top-down attention $M$ is initialized as all zeros $M_{0}$ and accumulated for all ground truth bounding boxes:

\begin{equation}
    M = M_{0} + \sum_{i=1}^{\left | G \right |}\sum_{j=1}^{\left | P \right |}  O_{i,j},  
\end{equation}
where $\left | G \right |$ is the number of ground truth bounding boxes. We do the same normalization as in Eq.~\ref{eq:bottomup}, using the top-down attention maps $M$ instead of the self-attention maps. Therefore, the top-down AFD objective is defined as:

\begin{equation}
    { L }^{\text{TD}}_{ \text{AFD} } (F^{B}) =  \frac { 1 }{ 2 } { \left\| \frac { { M }_{ i }^{ B } }{ \left\| { M }_{ i }^{ B } \right\|  } { F }^{ B }({ \mathbf{x} }_{ i })-\frac { {M }_{ i }^{ A } }{ \left\| { M }_{ i }^{ A } \right\|  } { F }^{ A }({ \mathbf{x} }_{ i }) \right\|  }^{ 2 }
\end{equation}

Top-down AFD focuses on distillation on foreground objects, which is crucial for object detection algorithms. To leverage both rich context information and foreground objects, The final AFD loss function is defined as:
\begin{equation}
    {L}_{\text{AFD}} (F^{B}) = {L}^{\text{BU}}_{\text{AFD}} + {L}^{\text{TD}}_{\text{AFD}} .
\end{equation}

\subsection{Adaptive Exemplar Sampling}
\label{sec:sampling}
In incremental learning for classification, it is common to store a small number of samples (``exemplars'') to avoid catastrophic forgetting~\cite{chaudhry2018riemannian,rebuffi2017icarl} while still maintaining training efficiency. However, it is unclear how effective the exemplars are for object detection. Thus, we analyze three different ways of using exemplars for object detection. 

Random sampling is effective for classification problems~\cite{chaudhry2018riemannian}, so we use it as a baseline in object detection. Since the number of bounding boxes can vary in each sample image, we propose \emph{hard sampling} as another baseline, which consists of images with the most bounding boxes as exemplars. However, for object detection datasets ({\em e.g.},  KITTI or Kitchen), the diversity of samples can be reduced by using hard sampling, as the images with more bounding boxes are more likely to be sampled from consecutive frames.

To avoid this drawback, we propose \emph{adaptive sampling}: For each category, we rank all $K$ samples in decreasing order by number of bounding boxes. Given a budget of $s$ exemplars per class with $K\gg s$ and $\eta\in\mathbb{N}$, we randomly sample $s$ exemplars from the ordered list (1, $\eta s$). We define three different sampling cases: when $\eta =1$, we select the samples with the most bounding boxes (hard sampling), when $\eta \geq \frac{K}{s}$ we sample among all images randomly (random sampling), and when $ 1 < \eta$ and $\eta < \frac{K}{s}$, it balances the number of bounding boxes and diversity of exemplars through the choice of $\eta$.  It is defined as:

\begin{equation}
\begin{cases}
\text{hard sampling}, & \text{if} \: \: \eta =1 \\
\text{random sampling}, & \text{if} \: \: \eta \geq \frac{K}{s} \\
\text{adaptive sampling}, & \text{otherwise} \\
\end{cases}
.
\end{equation}
In the experimental section, we compare three different sampling methods and show the superior performance of the adaptive sampling.

\section{Experiments}
\label{sec:exp}

\tabscenarios

We use a Pytorch~\cite{paszke2017automatic} implementation of Faster-RCNN~\cite{jjfaster2rcnn} with SE-ResNet-50~\cite{seresnet} pre-trained on ImageNet. We fix the first convolution layer, the first residual block, and all BN layers during training. This is common practise in detection~\cite{ren2015faster,wang2019towards}. We consider different combinations of the following datasets:
Watercolor~\cite{inoue2018cross}, Clipart~\cite{inoue2018cross}, Comic~\cite{inoue2018cross}, Kitchen~\cite{georgakis2016multiview}, KITTI~\cite{geiger2012we}, Pascal VOC~\cite{everingham2010pascal}, COCO~\cite{lin2014microsoft}. All train and test splits are the same as in ~\cite{wang2019towards}. Specifically, for Pascal VOC, we trained on VOC2007 and VOC2012 trainval set and tested on VOC2007 test set. For COCO, we trained on COCO 2014 valminusminival and tested on minival, to shorten the experimental period. The Pascal VOC mean average precision (mAP)  metric is used for evaluation. 

\minisection{Training details.} Following~\cite{jjfaster2rcnn}, we use the default parameters for Faster-RCNN with 4 anchor scales and 3 anchor ratios. Learning rate is set to 0.01 for 10 epochs and 0.001 for the last 2. For small datasets (Watercolor, Clipart, Comic) we use a batch size of 8 on 4 synchronized GPUs. For the other datasets, we use a batch size of 16. For all experiments, we set $\lambda=10^{-4}$ from Eq.~\ref{eq:overall}.

\minisection{Comparison methods.} We compare  AFD
with LwF Detection~\cite{shmelkov2017incremental}, Feature Distillation~\cite{romero2014fitnets}, Attention Distillation~\cite{zagoruyko2016paying}, EWC~\cite{kirkpatrick2017overcoming}, MAS~\cite{aljundi2018memory} and two other baselines (joint training and fine-tuning).  For LwF Detection, we follow~\cite{shmelkov2017incremental} to embed the loss function into our framework. For joint training, we train a detector using all training data at the same time, which can be seen as \emph{upper bound} for incremental learning. Fine-tuning is done by training all object detection tasks sequentially without any additional loss. Unless specified, the adaptive sampling is used for all methods using exemplars for fair comparison.

\subsection{Analysis of the  AFD on Three Scenarios}
\label{sec:domains}
In this section, we analyze the effect of forgetting on different scenarios without using any exemplars to understand which factor impacts forgetting the most.

 As discussed in Section~\ref{sec:fourscenarios}, we decompose two-task incremental learning into 4 scenarios with different domains and categories differences. We focus on the three non-trivial scenarios, as show in Table~\ref{tab:scenario}:  1)  KITTI and Kitchen $\mathcal{D}_{A}\!\neq\!\mathcal{D}_{B}$, $\mathcal{Y}_{A}\!\neq\!\mathcal{Y}_{B}$; 2) Watercolor and Comic ($\mathcal{D}_{A}\!\neq\!\mathcal{D}_{B}$, $\mathcal{Y}_{A}\!=\!\mathcal{Y}_{B}$); 3) VOC and COCO ($\mathcal{D}_{A}\!=\!\mathcal{D}_{B}$, $\mathcal{Y}_{A}\!\neq\!\mathcal{Y}_{B}$). 

One interesting observation we can see from the results is that domain gaps have less impact in forgetting. Specifically, in Watercolor and Comic, forgetting is less severe compared to other two scenarios by a large margin, even for just fine-tuning. By using our proposed AFD, we can achieve similar performance as joint training. The drops are just $2.2$ and $0.6$ for each direction, respectively. One possible explanation for this phenomenon is that, although there exists domain discrepancy between Watercolor and Comic, since the categories and general perspective of the objects remain the same, the backbone features are quite similar between the two tasks. Training the RPN and detection head for the new task also would not disturb the backbone features much due to same categories. This can be seen from the low forgetting of vanilla fine-tuning.

The other observation is that if we continually train on a new task in the same domain, with all the classes from the previous task, i.e., VOC to COCO, there is no forgetting for our proposed AFD. On the other hand, there is severe forgetting of up to $20\%$  mAP when continually train on a new task with less classes from the previous task, i.e., COCO to VOC. In this case, neither fine-tuning nor AFD can avoid the forgetting, though AFD still manages to improve over Fine-tuning from $20.1\!\rightarrow\!27.8$. Therefore, learning the RPN and detection head on tasks with more classes actually enhance the backbone features for the previous task, while learning on tasks with less classes can cause severe forgetting. In the latter case, the same feature maps that produce positive signal in the previous RPN and detection head could be negative in new heads. Thus the gradients propagated back to the backbone are significantly different, causing severe forgetting.

Unsurprisingly, for Kitchen and KITTI, where there are the vast differences in categories and domains, we observe severe forgetting. Fine-tuning forgets the first task almost completely while our proposed AFD can reduce the performance drops from $53.7\!\rightarrow\!36.6$ and $70.9\!\rightarrow\!20.5$, respectively. However, with just attention, it is hard to mitigate the forgetting effect even with the best distillation methods for this scenario. To solve the forgetting problem in the difficult cases for the last two scenarios, we explore the usage of exemplars in the following sections.

Overall, EWC and MAS performs worse than our proposed AFD in most cases, while MAS achieves good performance from COCO to VOC and Watercolor to Comic. 

\tabkittikitchen

\tabcomicwatercolor

\tabcocovoc

\subsection{Comparisons on Two-Task Setting}
In this section, we compare to all methods in three different scenarios for two-task setting. We have shown in previous section, for difficult cases, forgetting is severe. Keeping exemplars can further mitigate forgetting. We add ‘*' to indicate methods combining with exemplars.

\minisection{KITTI and Kitchen.} As shown in Table~\ref{tab:kit}, Fine-tuning suffers from catastrophic forgetting without exemplars, however with 100 exemplars we can see that Fine-tuning* performs quite well and obtains similar accuracy as LwF Detection*~\cite{shmelkov2017incremental}. Feature Distillation* is worse in both cases, while Attention Distillation* performs better by filtering out the noisy information. It is interesting that EWC* and MAS* outperforms other methods by integrating with exemplars. Our AFD* outperforms all other methods and baselines and is close to joint training with only 100 exemplars (less than 3\% of whole datasets).

\minisection{Comic and Watercolor.} In the scenario of domain shift when we have the same categories but different domains, we compare with the methods and baselines without using exemplars due to much less forgetting, as demonstrated in Table~\ref{tab:scenario}. As seen in Table~\ref{tab:comicw}, Attention Distillation and MAS achieve the best among baselines but they are worse than ours. our AFD achieves 46.6 mAP compared to Joint Training with 47.5 in average.  It shows that we can avoid forgetting in this scenario even without using any data from previous tasks, which is much efficient than Joint training. 

\minisection{COCO to VOC.} We present results on this relatively larger datasets in Table~\ref{tab:voc}. Since there is no forgetting from VOC to COCO (seen in Table~\ref{tab:scenario}), we only report results from COCO to VOC. COCO is a more challenging dataset compared to others because of the large amount of data and categories. Our AFD* surpasses the Attention Distillation* methods by $5.2$. EWC* and MAS* are the best baselines in this case, still our method is superior to them. However, there is still space for improvement until achieving similar performance as joint training. We also show some qualitative results on COCO before and after training on VOC (see Figure~\ref{fig:coco}). We can clearly see that Fine-tuning* with exemplars outputs less bounding boxes or has lower prediction confidence on COCO after learning on VOC. Our proposed AFD* closes the gap between fine-tuning and joint training by forgetting less and predicting more confident scores.

\figforgetting
\figvisualization 
\subsection{Comparisons on More Tasks}
\label{sec:moretasks}
\tabthree
\tabsampling
In this section, we show that our method is capable of extending to many tasks. Specifically we compare on two settings of learning three tasks and six tasks incrementally.

\minisection{On three tasks.} We show results in more realistic and challenging settings learning incrementally with three tasks. In Table~\ref{tab:three}, we perform a sequence of incremental training on KITTI, VOC, and Kitchen. We compare our method with a competitive baseline, as demonstrated in previous experiments. The performances are evaluated after training the last task. Our proposed approach clearly outperforms Fine-tuning w/ exemplars by a large margin (5.6 in average of three tasks). 

\minisection{On six tasks.} To further scale-up our incremental object detector, we conduct experiments on a sequence of six different datasets: VOC, Clipart, KITTI, Watercolor, Comic, and Kitchen. To illustrate the dynamic process of forgetting after training on each dataset, we show in Figure~\ref{fig:six} a forgetting matrix, which reflects the dynamic of forgetting after training on each task. The overall forgetting of VOC, Clipart, KITTI, Watercolor and Comic is calculated by adding all values on each row. It is 19.5 / 12.5 / 3.7 / 7.3 / 2.5 For our method and 27.1 / 16.3 / 7.2 / 8.4 / 4.6 for Fine-tuning w/ exemplars, respectively. It shows that our method scales well when it applies to many more tasks.

\subsection{Ablation Study}

\minisection{Ablation study for different sampling methods} As shown in Table~\ref{tab:sampling}, we compare with the different sampling strategies introduced in Section~\ref{sec:sampling}. All results are reported with the proposed AFD. Hard sampling performs better than random sampling from Kitchen to KITTI, but is much worse from KITTI to Kitchen. When looking at the selected samples, we found that many consecutive frames have a very similar scene, which reduces the diversity of samples. By choosing different values of $\eta$, we observe that adaptive sampling outperforms both random and hard sampling with $\eta=5$ being the best. 

\minisection{Ablation study for number of exemplars} Although we have shown that AFD can mitigate forgetting by a large margin, there is still a gap with Joint Training in the more challenging scenario (i.e. KITTI - Kitchen). Therefore, we experiment with exemplar sampling by keeping a small number of exemplars from the previous task. As shown in Figure~\ref{fig:exemplar}, it is notable that with only $100$ randomly sampled exemplars (less than 3\% fraction of data), Fine-tuning increases by ${\sim}50\%$ from Kitchen to KITTI, and by ${\sim}25\%$ from KITTI to Kitchen. Combined with our distillation methods, the accuracy from previous task gets closer to Joint Training. Performance improves as more exemplars are added. AFD is superior to both individual bottom-up and top-down attentions.

\figexemplars  

\section{Conclusion}

In this paper, we study the problem of multi-task incremental detection by decomposing the two-task case into three different scenarios. With the proposed attentive feature distillation (AFD) on the three non-trivial scenarios, we gain insights on how domain gaps and category differences impact forgetting. Specifically, in our experiments, domain gaps cause less forgetting, and the category differences are actually the bigger problem (See Section~\ref{sec:domains}). In these difficult cases, sampling a small number of exemplars can largely mitigate forgetting.

{\small
\bibliographystyle{ieee_fullname}
\bibliography{egbib}
}

\end{document}